# In silico generation of novel, drug-like chemical matter using the LSTM neural network

Peter Ertl[1], Richard Lewis[1], Eric Martin[2], Valery Polyakov[2]

1 Novartis Institutes for BioMedical Research, Novartis Campus, CH-4056 Basel, Switzerland
2 Novartis Institutes for BioMedical Research, 5300 Chiron Way, Emeryville, California 94608-2916, United States

Corresponding author: peter.ertl@novartis.com; web: peter-ertl.com

December 2017

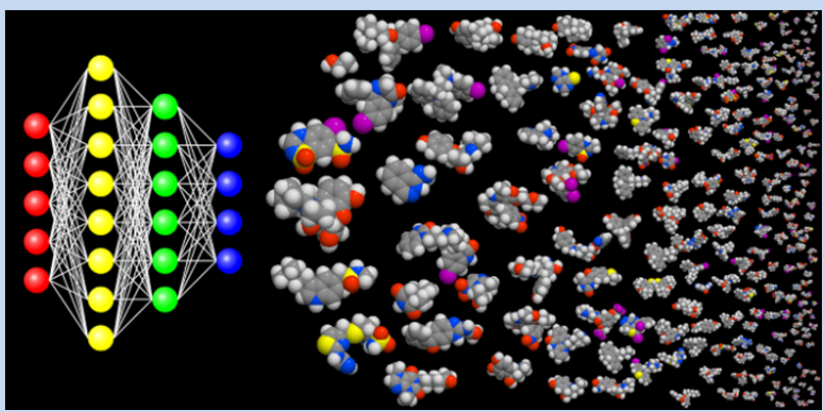

The exploration of novel chemical spaces is one of the most important tasks of cheminformatics when supporting the drug discovery process. Properly designed and trained deep neural networks can provide a viable alternative to brute-force de novo approaches or various other machine-learning techniques for generating novel drug-like molecules. In this article we present a method to generate molecules using a long short-term memory (LSTM) neural network and provide an analysis of the results, including a virtual screening test. Using the network one million drug-like molecules were generated in 2 hours. The molecules are novel, diverse (contain numerous novel chemotypes), have good physicochemical properties and have good synthetic accessibility, even though these qualities were not specific constraints. Although novel, their structural features and functional groups remain closely within the drug-like space defined by the bioactive molecules from ChEMBL. Virtual screening using the profile QSAR approach confirms that the potential of these novel molecules to show bioactivity is comparable to the ChEMBL set from which they were derived. The molecule generator written in Python used in this study is available on request.

## Introduction

The exploration of novel chemical spaces is one of the most important tasks of cheminformatics in supporting drug discovery. One way to address this problem is a "brute force" approach, trying to enumerate all reasonable molecules within certain size constraints. Due to the combinatorial explosion of the number of possible molecules to be generated with increasing size, this approach is currently possible only for relatively small molecules. An example of this approach is the GDB-17 database containing 166 billion molecules [1]. Full enumeration, however, of molecules with 25-30 atoms (the average size of drug molecules) is currently not possible. One needs to utilize various sophisticated cheminformatics algorithms to sample representative chemical space to generate novel molecules of interest, for example combination of drug-like fragments, evolutionary algorithms or particle swarm optimization. Review [2] provides a good overview of molecule generating methods.

Recent advances in neural networks has opened a new avenue for in silico generation of novel interesting molecules. Particularly the recently introduced long short-term memory (LSTM) networks [3] are well suited for this purpose. The inner working of the LSTM network is quite complex and a detailed description is beyond the scope of this article. Simply speaking, thanks to their sophisticated architecture, LSTM networks are able to "learn" characteristics of the objects they are trained on, and then answer questions about these objects and even generate new similar objects. LSTM networks have been used to support many new sophisticated technologies, including speech recognition, language translation, smart mobile phone assistants or self-driving cars [4]. Also numerous "lighter" applications that nicely document the power of the LSTM algorithm have been described. When trained on a set of Shakespeare's plays the network was able to create novel "Shakespeare" sonnets [5], or when trained on some classical piano music, to compose new songs [6]. The characteristics described above make the LSTM recurrent neural network also a good candidate for cheminformatics applications and particularly for generation of novel chemical matter.

The use of LSTM neural networks to generate novel molecules with desired properties is a hot topic and several academic as well as industrial groups are exploring this area. Bjerrum and Threlfall [7] used molecules from the ZINC database to train an LSTM network to generate novel molecules. Samples from these molecules have been checked using a retrosynthetic planning software and a retrosynthetic route could be identified for most of them. Use of several variations of training SMILES has been shown to provide better results than just a single canonical representation [8]. Gomez-Bombarelli *et al.* [9] trained an encoder-decoder network where the encoder was converting a molecule into a vector in a latent space, while the decoder was converting this vector back to the molecular structure. Operation in the continuous latent space allowed design of various drug-like molecules and light-emitting diodes. Olivecrona *et al.* [10] designed a network to generate molecules with desired properties, including analogues to a query structure and molecules with increased probability to be active again certain biological targets. Segler *at al.* [11] used molecules generated by an LSTM recurrent network to design several focused libraries targeting the serotonin receptor and several antibacterial targets. LSTM networks have also been successfully applied to retrosynthetic reaction prediction by Liu *et al.* [12], who trained a sequence-to-sequence reaction prediction network on 50,000 reactions from patents. The method performed comparably with a rule-based expert system. Yang *et al.* [13] combined a Monte Carlo tree search and a recurrent neural network to generate molecules with desired values of octanol-water partition coefficient and good synthesizability. Recently Gupta *et al.* [14] presented an application of recurrent networks for de novo drug design and fragment-based drug discovery. During the submission process of this manuscript study of Popova *et al.* [24] appeared, describing computational strategy for *de novo* drug design based on deep and reinforced learning.

## Methodology

### Neural network architecture

The open source high-level framework Keras [15] with the TensorFlow numerical backend engine [16] was used for this project. These popular software tools, written in Python, make development and application of deep neural networks relatively easy, allowing one to focus on the scientific aspects of the problem rather than on the details of numerical implementation. The major challenge in the application of LSTM networks is the design of a proper network architecture. The optimal combination of various network parameters (types and sizes of the network layers, activation and optimization functions, setting the learning and dropout rate and optimal way of encoding the training data into vectorised form required by the network) requires some trial and error. There are no general rules for designing the optimal architecture, one has to experiment and try different architectures and parameters. After several experiments the network architecture shown on Fig. 1 was selected for our study.

```
Using TensorFlow backend.
corpus length: 23664668
total chars: 23
nb sequences: 7888210
Vectorization...
Build model...
_________________________________________________________________
Layer (type)                 Output Shape              Param #
=================================================================
lstm_1 (LSTM)                (None, 40, 128)           77824
_________________________________________________________________
lstm_2 (LSTM)                (None, 64)                49408
_________________________________________________________________
dropout_1 (Dropout)          (None, 64)                0
_________________________________________________________________
dense_1 (Dense)              (None, 23)                1495
_________________________________________________________________
activation_1 (Activation)    (None, 23)                0
=================================================================
Total params: 128,727.0
Trainable params: 128,727.0
Non-trainable params: 0.0
_________________________________________________________________
```

Figure 1. The network architecture used in this study.

The network consists of double LSTM layers (that actually learn the SMILES grammar) a dropout layer and an output dense layer with 23 neurons (the number of characters we want to predict probability for) with the softmax activation function. The default RMSProp optimizer was used for the training. During training, it was necessary to continually decrease the learning rate from 0.01 to 0.0002. When using the trained network to generate structures of novel molecules a certain level of randomness (as suggested in [17]) was added to get more diverse structures that do not too closely reproduce the molecules in the training set. It is worth restating that this architecture may not be optimal, but it is sufficient to produce good results (see below).

## Training data

SMILES codes of 509,000 bioactive molecules (with activity on any target below 10 µm) from the ChEMBL database [18] were used to train the network. The SMILES codes were filtered and slightly modified to make the training easier. Only molecules containing the organic subset of atoms (H, C, N, O, S, P, F, Cl, Br, I) were used. Charged molecules were discarded and stereo information was not considered. Only molecules with up to 5 ring closures were retained in the training set. The atomic codes consisting of multiple characters (Cl, Br and [nH] in this case) were replaced by single character codes (L, R and A, respectively). After these modification the training corpus consisted of 23 characters (atomic symbols with upper- and lowercase symbols representing the aliphatic and aromatic atoms, the = and # bond symbols, parentheses, numbers 1 to 5 representing ring closures and a newline). The whole training corpus contained 23,664,668 characters. During the training process, sequences of 40 characters from the corpus were fed to the network together with a character following the sequence. The goal of the training was to learn this character from a preceding sequence. During the process the LSTM network learned the SMILES grammar and learned which character should most probably follow after certain group of characters. The Python code used to generate molecules in this study is available for download as an additional file.

## Results

New SMILES codes were generated character-by-character using the trained LSTM neural network. Not all SMILES codes generated were fully grammatically correct. Therefore, as the first step we checked whether all brackets and ring closure numbers are paired. This test could be performed relatively fast as a simple text query without necessity to parse the SMILES. 54% of generated SMILES codes could be discarded in this test. An additional 14% were discarded during the full chemical parsing using RDKit [19] due to unrealistic aromatic systems or incorrect atomic valences. At the end, 32% of the generated SMILES strings led to correct molecules. The generation of one million correct molecules required less than 2 hours on 300 CPUs. Switching to GPUs and increasing the number of processors would allow generation of hundreds of millions or even billions of molecules over a weekend.

The most important question we wanted to answer concerned the quality of generated molecules. Out of one million molecules generated, only 2774 were identical with molecules in the training ChEMBL set. The similarity distribution of the generated molecules to the ChEMBL compounds is shown in Fig. 2. The standard RDKit similarity based on Morgan fingerprints and Tanimoto coefficient was used [19]. One million generated structures contained 627 thousand unique scaffolds (the scaffold is part of the molecule after removal of non-ring substituent, for molecules without rings it is the largest chain). For comparison the ChEMBL training set contained 172 thousand scaffolds. The overlap between the two sets was only 18 thousand scaffolds. These data illustrate that our protocol generates novel molecules and does not mirror the training set too closely. The generated molecules also have good properties. Calculated common molecular descriptors (molecular weight, logP and TPSA [20]) shown in Figure 3 are very similar to those of the ChEMBL set.

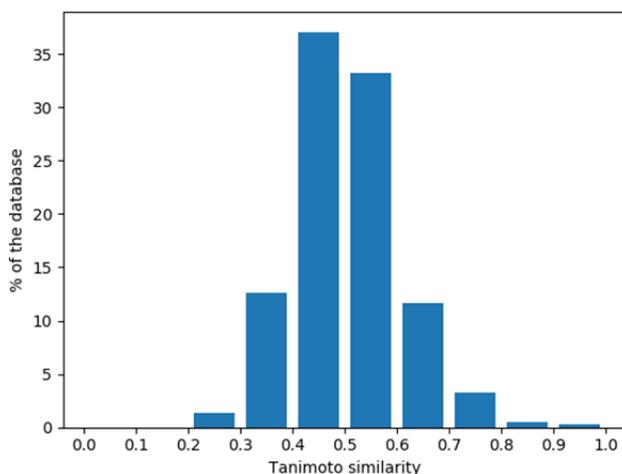

Figure 2. Similarity of molecules generated by the LSTM network to the ChEMBL training molecules

In Table 1, counts of the most important substructure features in the ChEMBL training set and the molecules generated by the LSTM network are shown. The table also contains a 3rd column showing statistics for molecules generated by a simple naive SMILES generator (denoted as baseline in the Table 1). By this analysis we wanted to see whether the "intelligence" of the deep LSTM neural network is necessary to generate valid molecules, or whether a simple SMILES generator using only the syntax of the SMILES language, and some basic statistics derived from a reference database of known structures is sufficient. This tool analyzed the training set to provide the frequencies of occurrence of atom types and the average number of symbols between pairing operations (ring closures and brackets). These were used as roulette wheel probabilities. SMILES generation started by picking an element according to frequency, and the string was then filled up character by character with symbols, rings and branches. Aromatic elements could only be added between ring closures. At the end, the SMILES string was parsed using RDKit to see if it represents a valid molecule.

This analysis clearly shows that the structures generated by the LSTM neural network have a distribution of substructure features very similar to those of the ChEMBL molecules, while the naive SMILES generator using only knowledge of the statistical distribution of characters in the SMILES strings and basic SMILES syntactic rules generates "ugly" molecules, mostly macrocycles and practically no fused aromatic systems, and is not able to reproduce the chemical space of the ChEMBL molecules.

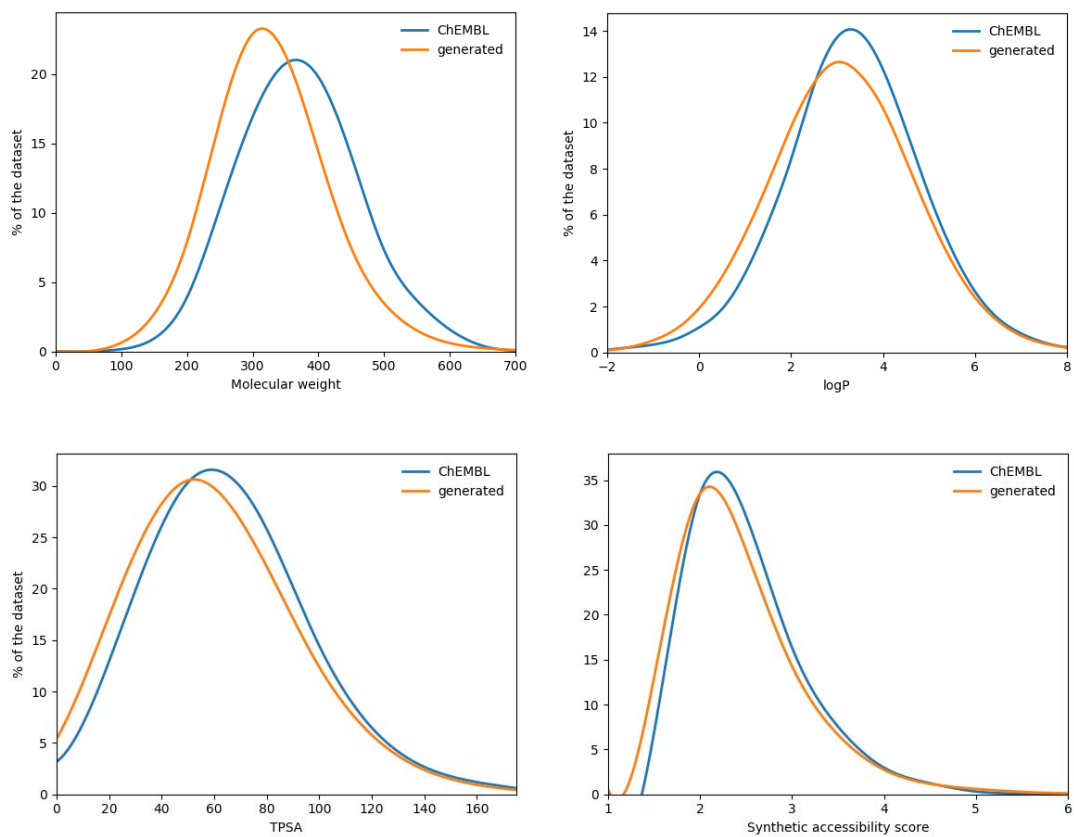

Figure 3. Calculated molecular properties and synthetic accessibility score for the ChEMBL training set and generated molecules.

Table 1
Comparison of substructure features between the ChEMBL training set, molecules generated by the LSTM network and by the naive SMILES generator.

| Feature | ChEMBL | generated | baseline |
|---|---|---|---|
| no rings | 0.4 | 0.4 | 0.1 |
| 1 ring | 2.8 | 4.3 | 13.2 |
| 2 rings | 14.8 | 23.1 | 17.7 |
| 3 rings | 32.2 | 43.5 | 27.3 |
| 4rings | 32.7 | 23.9 | 25.2 |
| >4 rings | 17.2 | 4.8 | 16.5 |
| fused arom. rings | 38.8 | 30.9 | 0.2 |
| large rings (> 8) | 0.4 | 1.8 | 75.9 |
| spiro rings | 1.9 | 0.6 | 0.6 |
| without N,O,S | 0.0 | 0.2 | 2.6 |
| contains N | 96.5 | 96.1 | 92.3 |
| contains O | 93.0 | 92.0 | 85.5 |
| contains S | 35.6 | 27.9 | 39.6 |
| contains halogen | 40.7 | 38.8 | 49.4 |

Comparison of the most common functional groups [21] in both sets shown in Figure 4 indicates that both the ChEMBL molecules and the structures generated by the LSTM neural network have very similar distributions of functional groups, again confirming that both sets occupy the same chemical space.

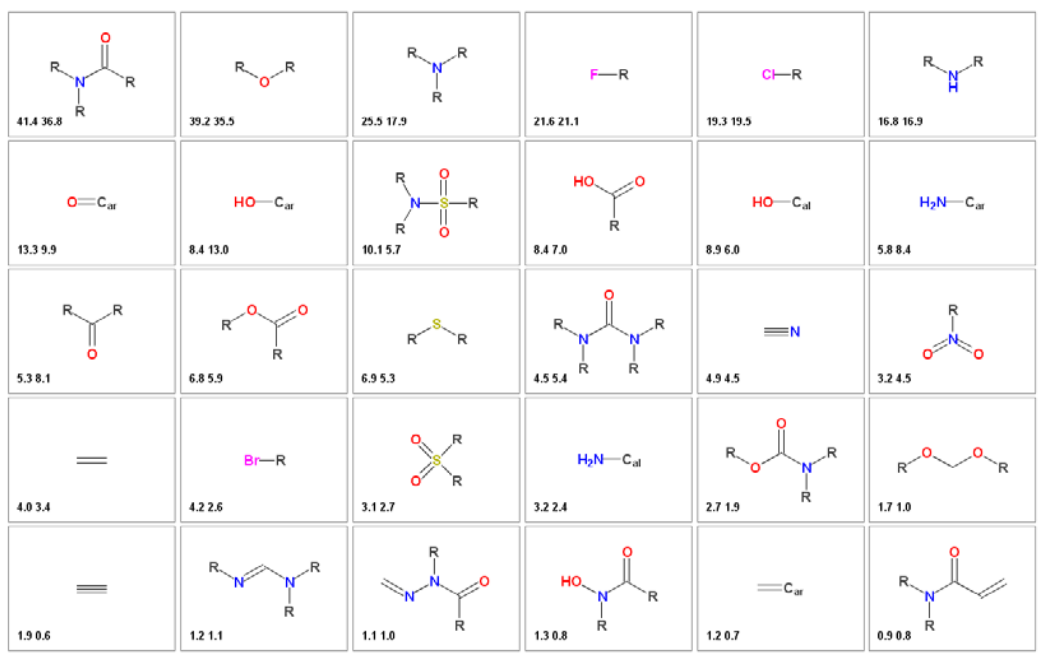

Figure 4. Comparison of the percentage of occurrences of the important functional groups in the ChEMBL training set (first number) and the generated molecules (second number).

For both training and generated molecule sets, we also calculated the synthetic accessibility score [22], a measure of how easy it might be to synthesize the molecule. The distribution of the score (shown as the last graph in the Figure 2) is practically identical for both sets, indicating that the generated structures are of similar complexity as the drug-like molecules in the ChEMBL database.

In summary, the generated structures are novel, having very little overlap with the training set, are diverse, containing large number of novel scaffolds, and their properties are in the same range as the drug-like ChEMBL molecules. The substructure analysis and the distribution of functional groups in the newly generated molecules indicates that the LSTM network is able to learn the drug-like structural features of the training set well and introduce them into the newly generated structures.

## Virtual screening

The likelihood of the novel molecules showing bioactivity profiles comparable to the actual ChEMBL molecules was assessed by virtual screening. Profile-QSAR multi-target virtual screening models have been previously reported for

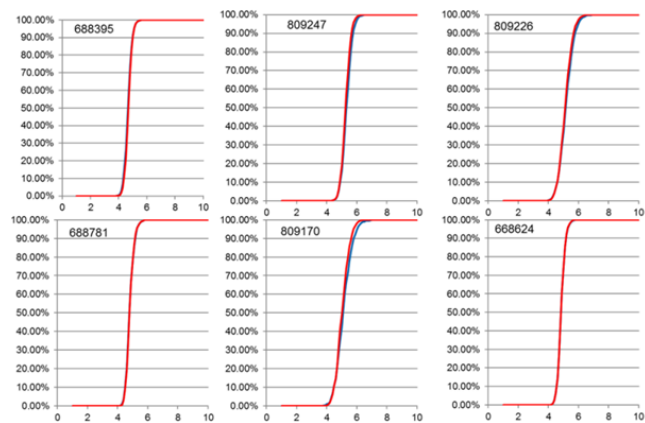

Figure 5. Comparisons of predicted pIC$_{50}$ cumulative frequency distributions between actual (blue) and generated (red) compounds for the 6 best ChEMBL kinase profile-QSAR models. All models had correlation between prediction and experiment of $R^2 > 0.75$ for the very challenging "realistic" test sets.

159 ChEMBL kinase assays [23]. Many of these models have very good prediction ability. Predictions from the top six models correlated with experiment with $R^2 > 0.75$ on held-out test sets of the most novel possible compounds.

The activity of the actual ChEMBL structures and generated structures were predicted with these 6 models. Figure 5 overlays the cumulative frequency distributions (CFD) of predicted $pIC_{50}$ for the actual and generated ChEMBL compounds for these top 6 models. The overlays are almost perfect indicating that the generated compounds should have similar likelihoods of biological activity as the actual ChEMBL compounds.

The Kolmogorov-Smirnov (KS) test quantitatively compares CFD curves by examining the largest vertical distance between the curves. The test was applied to random samplings of 1000 compounds from each model. Tee table 2 shows the results of the KS test along with the means and standard deviations for the activity distributions of the actual and generated compounds for the 6 models. For 4 of the 6 models, the KS test allows the null hypothesis that the distributions are the same with 95% confidence. For 2 of the assays, 809247 and 809170, the KS test rejects the null hypothesis that the distributions are the same with 95% confidence, but even in the worst case of 809170 the largest vertical distance is <10%. Within each assay, the pairs of means and standard deviations for the actual and generated distributions coincide very closely, much closer than the variation in the means and standard deviations between assays, indicating that the generated and actual compounds would have similar activity profiles.

Table 2. Profile-QSAR Predicted Activity Distributions for 6 ChEMBL Assays

| Assay[1] | KS D[2] | Differ?[3] | Mean real[4] | Mean gen[5] | Stdev real[6] | Stdev gen[7] |
|---|---|---|---|---|---|---|
| 688395 | 6.01% | no | 4.66 | 4.69 | 0.25 | 0.24 |
| 668624 | 3.60% | no | 4.86 | 4.86 | 0.25 | 0.24 |
| 809226 | 9.90% | yes | 5.33 | 5.26 | 0.34 | 0.30 |
| 809226 | 4.30% | no | 5.18 | 5.13 | 0.47 | 0.43 |
| 688781 | 2.20% | no | 4.83 | 4.82 | 0.26 | 0.25 |
| 809170 | 8.70% | yes | 5.12 | 5.07 | 0.51 | 0.46 |

[1] ChEMBL assay ID
[2] Kolmogorov Smirnov D statistic which is the maximum vertical distance between cumulative frequency distribution curves.
[3] Critical value of D at alpha=0.05 is 6.04%. If D>6.04% the distributions are different with 95% confidence.
[4] Mean predicted activity for real ChEMBL compounds
[5] Mean predicted activity for generated ChEMBL compounds
[6] Standard deviation of predicted activities for real ChEMBL compounds
[7] Standard deviation of predicted activities for generated ChEMBL compounds

## Conclusions

The LSTM neural networks have been used to generate novel molecules based on a training set of 550 thousand drug-like molecules from the ChEMBL database. The generated structures are novel, diverse, have good properties and synthetic accessibility. Although novel, the molecules occupy the same area of chemical space as the known bioactive molecules. Virtual screening indicates that the generated molecules should have comparable probability of activity on various targets.

The performance of the algorithm is sufficient to generated hundreds of millions, even billions of new drug-like molecules. Novel structures generated in this way may be used in virtual screening and in exploring novel interesting areas of the chemical universe.

The Python code used to train the LSTM network and to generate SMILES codes of new molecules described in this article is available from the author (PE) on request.

## Author's contributions

PE designed the network and performed the generation of the molecules and their cheminformatics analysis, RL wrote the naive SMILES generator, EM and VP performed the virtual screening and analyzed its results. All the authors wrote the article together.

## Acknowledgements

We want to thank the following Novartis colleagues that helped by suggestions or by discussing the results: Niko Fechner, Brian Kelley, Stephan Reiling, Bernd Rohde, Gianluca Santarossa, Nadine Schneider, Ansgar Schuffenhauer, Lingling Shen, Finton Sirockin, Clayton Springer, Nik Stiefl and Wolfgang Zipfel.